\documentclass[runningheads]{llncs}
\usepackage{graphicx}
\usepackage{amsmath,amssymb} % define this before the line numbering.
\usepackage{bm}
\usepackage{bbm}
\usepackage{color}
\usepackage{sistyle}
\begin{document}
\title{Towards Good Practices for Multi-modal Fusion in Large-scale Video Classification} 
% Replace with your title

\titlerunning{Multi-modal Fusion in Large-scale Video Classification}
% Replace with a meaningful short version of your title
%
\author{Jinlai Liu \and
Zehuan Yuan \and
Changhu Wang }
%
%Please write out author names in full in the paper, i.e. full given and family names. 
%If any authors have names that can be parsed into FirstName LastName in multiple ways, please include the correct parsing, in a comment to the volume editors:
%\index{Lastnames, Firstnames}
%(Do not uncomment it, because you may introduce extra index items if you do that, we will use scripts for introducing index entries...)
\authorrunning{Jinlai Liu, Zehuan Yuan, Changhu Wang}
% Replace with shorter version of the author list. If there are more authors than fits a line, please use A. Author et al.
%

% \email{lncs@springer.com}

\institute{Bytedance AI Lab  \\
\email{\{liujinlai.licio,yuanzehuan,wangchanghu\}@bytedance.com}
}

\maketitle

\begin{abstract}
Leveraging both visual frames and audio has been experimentally proven effective to improve large-scale video classification. Previous research on video classification mainly focuses on the analysis of visual content among extracted video frames and their temporal feature aggregation. In contrast, multimodal data fusion is achieved by simple operators like average and concatenation. Inspired by the success of bilinear pooling in the visual and language fusion, we introduce multi-modal factorized bilinear pooling (MFB) to fuse visual and audio representations. We combine MFB with different video-level features and explore its effectiveness in video classification. Experimental results on the challenging Youtube-8M v2 dataset demonstrate that MFB significantly outperforms simple fusion methods in large-scale video classification.

\end{abstract}

\section{Introduction}

Along with the dramatic increase in video applications and production, better video understanding %and recognizing 
techniques are urgently needed. As one of the fundamental video understanding tasks, multi-label video classification has attracted increasing attentions in both computer vision and machine learning communities. Video classification requires a system to recognize all involved objects, actions and even events in any video based on its available multimodal data such as  visual frames and audio. 

%existing research work mainly employs deep neural network for feature extraction, feature aggregation and multi-label prediction. 
As deep learning has obtained a remarkable success in image classification ~\cite{Krizhevsky2012ImageNet,He2016Deep,Szegedy2016Rethinking}, action recognition~\cite{simonyan2014two,wang2016temporal,Girdhar2017ActionVLAD} and speech recognition~\cite{hinton2012deep,vgg_audio}, video classification also benefit a lot from these powerful image, snippet, and audio representations.
Since videos are composed of continuous frames, aggregating frame or snippet features into video-level representation also plays an important role during recognition process. 
%For feature extraction, we usually utilize deep convolutional nerual networks(CNNs)~\cite{Krizhevsky2012ImageNet}~\cite{He2016Deep}~\cite{Szegedy2016Rethinking}~\ pretrained on large-scale images datasets~\cite{deng2009imagenet}.
Besides the direct aggregations such as temporal average or maxpooling, a few sophisticated temporal aggregation techniques are also proposed. For example, Abu-El-Haija {\it et al.}~\cite{Abu2016YouTube} proposes deep Bag-of-Frames pooling (DBoF) to sparsely aggregate frame features by ReLU activation. % ~\cite{arandjelovic2016netvlad} reproduce the VLAD encoding~\cite{J2010Aggregating} to enable end-to-end train cluster weights. 
On the other hand, recurrent neural networks such as long short-term memeory (LSTM)~\cite{hochreiter1997long} and gated recurrent unit (GRU)~\cite{cho2014properties}) are applied to model temporal dynamics along frames.
%\{..clustering based pooling such as NetVLAD~\cite{arandjelovic2016netvlad}, DBoF~\cite{Abu2016YouTube} and temporal memory network using in sequential modeling such as some variants of recurrent network(long short-term memory (LSTM)~\cite{hochreiter1997long}, gated recurrent unit(GRU)~\cite{cho2014properties})... \}
%We can see that these techniques are mainly borrowed or inspired by modeling in single modal processing. 

Although much progress has been made in generating video-level visual representations, few work lies on integrating multimodal data which can supplement with each other and further reduce the ambiguity of visual information. Therefore, 
%However, video is the most informative, multimodal data source. 
%Video is content-rich with visual, audio, caption, etc. In most recent research work, the input of video classification model is basically raw RGB or its optical flow. The effect of multi-modal inputs and fusion techniques have rarely been
 %tried and verified. 
%
developing deep and fine-grained multimodal fusion techniques could be a key ingredient towards practical video classification systems. In this paper, we take the first step by introducing multi-modal bilinear factorized pooling into video classification, which has been extensively adopted in visual question answering ~\cite{Gao2015Compact,kim2016hadamard,yu2017mfb}. We select three popular video-level representations, i.e, Average pooling,  NetVLAD~\cite{arandjelovic2016netvlad} and DBoF~\cite{Abu2016YouTube}, to validate its effectiveness. Experimental results indicate that video classification can achieve a significant performance boost by leveraging the new pooling mechanism over video and audio features. 
%We further verify the effectiveness of multi-modal factorized bilinear pooling is not due to the increase the parameter size.
In summary, our contributions are twofold:
\begin{itemize}

\item[$\bullet$] We first introduce multi-modal factorized bilinear pooling to integrate visual information and audio in large-scale video classification.

\item[$\bullet$] We experimentally demonstrate that multi-modal factorized pooling significantly outperforms simple fusion methods over several video-level features.

\end{itemize}

\section{Related Work}
% This work is related to previous methods for video classification task, especially the temporal modeling, and then shows how the multi-modal bilinear fusion apply to visual and language multi-modal learning. 

\subsection{Video Classification}
Large-scale datasets ~\cite{deng2009imagenet,caba2015activitynet} play a crucial role for deep neural network learning. In terms of video classification, Google recently releases the updated Youtube-8M dataset~\cite{Abu2016YouTube} with 8 millions videos totally. For each video, only visual and audio representations of multiple frames are made public.  
% \subsection{Temporal Modeling}
%Until recently, 
The approaches for video classification roughly follow two main branches. On the one hand, several architectures are introduced to extract powerful frame or snippet representations similar to image classification. Simonyan and  Zisserman {\it et al.} first introduces deep convolutional neural networks to video action classification by performing frame-level classification ~\cite{simonyan2014two}. In order to include more temporal information, 3D convolutional neural network and several variants ~\cite{tran2018closer,qiu2017learning,carreira2017quo} are proposed to generate representations of short snippets. The final video predictions can be  estimated by late fusion or early fusion.  On the other hand, researchers also direct their eyes to how to model long-term temporal dynamics when frame-level or snippet-level representation available. %Compared with single image spatial modeling, video is a more complex information carrier. So meeting with the needs of temporal reasoning is the primary research point before. Temporal modeling approaches focus on how to efficiently aggregate and extract label-aware feature from long consecutive frame-level features . 

Commonly used methods to model long-term temporal dynamics are various variants of Bag of Visual Words (BoVW) including Vector of Locally Aggregated Descriptors (VLAD)~\cite{J2010Aggregating}, Fisher Vector (FV)~\cite{Perronnin2007Fisher} and so on. But these handcrafted descriptors cannot be finetuned for the target task. Therefore, an end-to-end trainable NetVLAD~\cite{arandjelovic2016netvlad} was proposed where a novel VLAD layer was plugged into a backbone convolutional neural network. Girdhar {\it et al.} proposed ActionVLAD that performs spatio-temporal learnable aggregation for video action classification. On the other hand, temporal models such as LSTM and GRU, are also widely used to aggregate frame-level features into a single representation due to its capability of capturing the temporal structures of videos.

\subsection{Multimodal Learning}

A simple attempt to integrate multimodal data is performing average or concatenation before input to final predictions. %ing visual and  feature in frame-level(Early Concat) or video-level(Late Concat). 
However, more fine-grained multimodal fusion models like bilinear pooling operations have been extensively explored and validated in visual and language multimodal learning. Lots of work has focused on addressing the huge number of model parameters and high computation cost in bilinear pooling. Multi-modal compact bilinear (MCB)~\cite{Gao2015Compact} was proposed to employ tensor sketch algorithm to reduce the computation cost and the amount of parameters. %effectively and simultaneously reduce the computation cost and the amount of parameters by a tensor sketch algorithm. 
Later, to overcome the high memory usage in MCB, multi-modal low-rank bilinear pooling (MLB)~\cite{kim2016hadamard} adopted Hadamard product to combine cross-modal feature vectors. Furthermore,  multimodal tucker fusion (Mutan)~\cite{ben2017mutan} and multi-modal bilinear factorized bilinear pooling  (MFB)~\cite{yu2017mfb} were proposed to address rather huge dimensionality and boost training.

In the paper, inspired by the success of MFB in visual and language fusion, we apply  MFB~\cite{yu2017mfb} into the video classification task by combining available visual and audio representations. The most related work to us is probably~\cite{Miech2017Learnable} which tried multi-modal compact bilinear pooling approach~\cite{Gao2015Compact} in large-video video classification but failed to fit training data.

\begin{figure}[t]
\centering
\includegraphics[height=4.8cm]{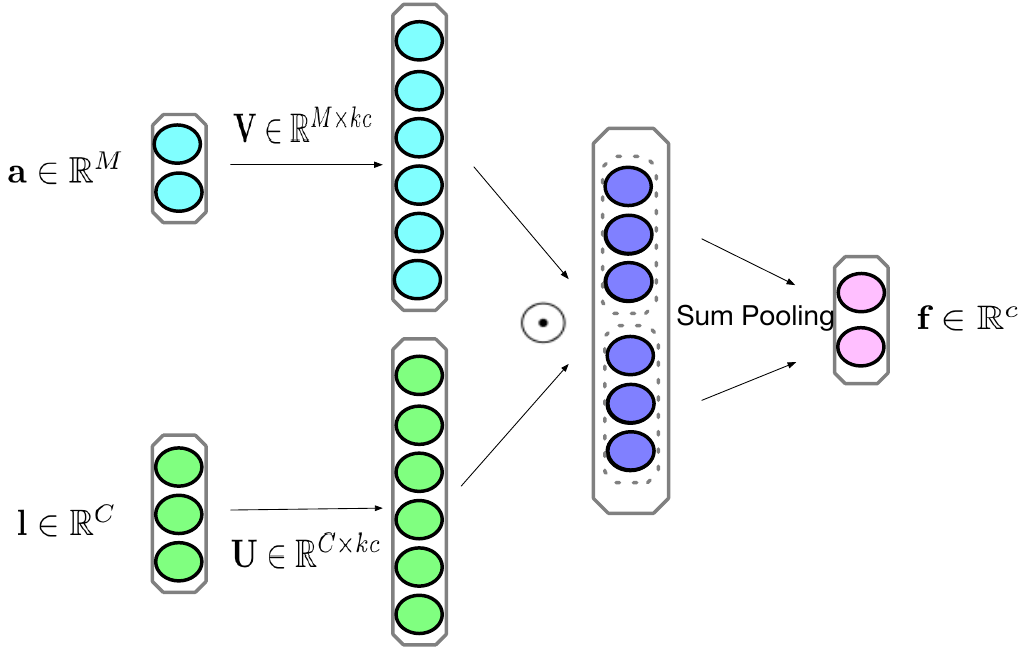}
\caption{The architecture of multi-modal factorized bilinear pooling.}
\label{fig:mfb_module}
\end{figure}

\section{Multi-modal Bilinear Fusion}
\label{sec:model}

We apply multi-modal factorized bilinear pooling over video-level visual and audio features. However in practice, only frame-level or snippet-level representations are available. Therefore as mentioned above, three methods are exploited to aggregate frame-level features into a single video feature. In this section,  we firstly review the MFB module and temporal aggregation models and then present our classification framework. 

\subsection{Multi-modal Factorized Bilinear Pooling}

%Bilinear models provide an appealing network structure for mixing and merging information from different modals.  
For any video, let $\textbf{l} \in \mathbb{R}^{C}$ and  $\textbf{a} \in \mathbb{R}^{M}$ denote its visual feature and audio feature, respectively. $M$ and $C$ are their corresponding feature dimensions. Then the output of MFB over $\textbf{l}$ and $\textbf{a}$ is a new vector $\textbf{f}$ with the $i$-th element formulated as
\begin{equation}
    f_{i} = \textbf{l}^{T}\textbf{W}_{i}\textbf{a},
\end{equation}
where $\textbf{W}_i \in \mathbb{R}^{C \times M}$. %The bias terms are omitted without loss of generality.
In order to reduce the number of parameters and the rank of weight matrix $\textbf{W}_i$, a novel low-rank bilinear model is proposed in~\cite{Pirsiavash2009Bilinear}. Specifically, $\textbf{W}_i$ is decomposed as the multiplication of two low-rank matrices $\textbf{U}_{i}$ and $\textbf{V}_{i}$, where $\textbf{U}_{i} \in \mathbb{R}^{C \times k}$ and $\textbf{V}_{i} \in \mathbb{R}^{M \times k}$. $k$ is a predefined constant to control rank. Therefore, 
\begin{equation} 
   \textbf{f}_{i} = \textbf{l}^{T}\textbf{W}_{i}\textbf{a} = \textbf{l}^{T}\textbf{U}_i\textbf{V}_i^T\textbf{a} = \mathbbm{1}^{T}(\textbf{U}_{i}^T\textbf{l})  \odot (\textbf{V}_{i}^T\textbf{a}),
\end{equation}
where $\mathbbm{1} \in \mathbb{R}^k$ is an all-one vector and $\odot$ denotes Hadamard product. It is worthy noting that $\mathbbm{1}$ is essentially a sum pooling operator as shown in Fig.~\ref{fig:mfb_module}. We follow the same normalization mechanism as in~\cite{Pirsiavash2009Bilinear} except that the power normalization layer is replaced with a ReLU layer in our MFB module implementation.

%For low-rank bilinear pooling model, a pooling matrix $P$ is introduced:
%\begin{equation}
 %   F = P^{T}(U_{v}^T\hat{V}) \odot (U_{a}^T\hat{A}),
%\end{equation}
%where $P \in \mathbb{R}^{k \times f}$, $U_{v} \in \mathbb{R}^{N \times k}$, $U_{a} \in \mathbb{R}^{M \times k}$. 
%Here we redefine $U_{v}$ and $U_{a}$ to be two-dimensional tensors% by introducing $P$ for a vector output $F \in \mathbb{R}^{f}$, 
%largely reducing the number of weights. $P$ plays the same role as the SumPooling operation in the MFB module shown in Fig.~\ref{fig:mfb_module}. 

\subsection{Temporal Aggregation Model}
In order to validate the general effectiveness of MFB in video classification, we experiment with video-level visual and audio features of three kinds obtained by average pooling, DBoF and NetVLAD over respective frame-level features. Let $\textbf{L}\in \mathbb{R}^{N\times C}$  and  $\textbf{A}\in \mathbb{R}^{N\times M}$ denote frame-level visual and audio features for a given video with $N$ frames, respectively. In our experiment, $C=1024$ and $M=128$.  $\textbf{L}$ and  $\textbf{A}$ are processed separately for each  pooling mechanism.
\subsubsection{Average Pooling (Avgpooling):} %Below we introduce three temporal aggregation modules in detail. , 
The average pooling layer is simply averaging features across $N$ frames, that is,
\begin{equation}
    \textbf{l} = \frac{1}{N} \sum_{i=1}^{n} \textbf{L}_i,  \textbf{a} = \frac{1}{N} \sum_{i=1}^{n} \textbf{A}_i.
\end{equation}

\subsubsection{DBoF:} Deep Bag-of-Frames pooling extends the popular bag-of-words representations for video classification~\cite{Ng2015Beyond,Wang2009Evaluation} and is firstly proposed in~\cite{Abu2016YouTube}. Specifically, the feature of each frame is first fed into a fully connected layer(fc) to increase dimension, %, as denoted in Eq.~\ref{eq:DBoF}:
% \begin{equation}
%      \label{eq:DBoF}
%     \hat{Vi} = ReLU(V_iW_m)
% \end{equation}
%where $W_m \in \mathbb{R}^{D \times M}$, and $W_m$ are shared across the $n$ input frames. Typically, with $M > D$, the input features are projected onto a higher dimensional space. 
%Along with the ReLU activations, this leads to a sparse coding of input features in the M-dimensional space.
Max pooling is then used to aggregate these high-dimensional frame-level features into a fixed-length representation. Following~\cite{Abu2016YouTube},  a rectified linear unit (RELU) and  batch normalization layer (BN) is used to increase non-linearity and keep training stable.

\subsubsection{NetVLAD:} The NetVLAD~\cite{arandjelovic2016netvlad,Girdhar2017ActionVLAD} employed the VLAD encoding~\cite{J2010Aggregating} in deep convolutional neural networks. The whole architecture can be trained in an end-to-end way. Compared to  VLAD encoding,  the parameters of clusters are learned by backpropagation instead of k-means clustering. %~\cite{Girdhar2017ActionVLAD} has implemented it on video action recognition task. 
Assuming $K$ clusters are used during training, NetVLAD assigns any descriptor $\textbf{h}_i$ in $\textbf{L}$ or $\textbf{A}$  to the cluster $k$ by a  soft assignment weight  

\begin{equation}
    \alpha_k(\textbf{h}_i) = \frac{\textbf{w}_{k}^T \textbf{h}_i + b_k}{\sum_{k'=1}^{K}e^{\textbf{w}_{k'}^T \textbf{h}_i + b_{k'}}}\textrm{,} 
\end{equation}

where $\textbf{w}_{k'}$ and $\textbf{b}_{k'}$ are trainable cluster weights. %That is the soft assignment $\alpha_k(v_i)$ of descriptor $v_i$ to cluster $k$ measures on a scale from 0 to 1 how close the descriptor $x_i$ is to cluster $k$. 
Compared to  the hard assignment, $\alpha_k(\cdot)$ measures the distance between descriptors with the cluster $k$ and thus maintains more information. % $\alpha_k(v_i)$ would be equal to $1$ if $v_i$ closest cluster is cluster $k$ and $0$ otherwise. 
%For the rest of the paper,  $\alpha_k(v_i)$ will define soft assignment of descriptor $v_i$ to cluster $k$. 
With all assignments for descriptors, the final NetVLAD representation is a weighted sum of residuals relative to each cluster. For the cluster $k$:
\begin{equation}
    VLAD[k] = \sum_{i=1}^{N} \alpha_k(\textbf{h}_i) (\textbf{h}_{i} - \textbf{c}_k) ,
\end{equation}
where $\textbf{c}_k$ corresponds to the learnable center of the $k$-th cluster. %computes the weighted sum of residuals $v_i - c_k$ of descriptors $v_i$ from learnable anchor point $c_k$ in cluster $k$.

% \subsection{Mixture of Experts(MoE) as the Classifier}

% The Mixture of Experts~\cite{Jordan1994Hierarchical}  classifier layer consists of  $m$ ``expert networks''. It obtains global multimodal representation $F \in \mathbb{R}^{f}$  as input and predicts a distribution of $c$ classes finally.
% \begin{equation}
%     G_{i} = FW_{g, i} + \lambda ||W_{g,i}||_{2},
% \end{equation}
% \begin{equation}
%     E_{i} = FW_{e, i} + \lambda ||W_{e,i}||_{2},
% \end{equation}
% \begin{equation}
%     O = \sum_{i=1}^{m}softmax(G_i) \odot sigmoid(E_i),
% \end{equation}
% where $W_{g,i}, W_{e,i} \in \mathbb{R}^{f \times c}, i \in \{1, ..., m\}$, $O \in \mathbb{R}^{c}$. To prevent overfitting, we also add L2 regularization, $\lambda$ is the L2 penalty with default value of $1e-6$. All of our models are trained with 2-mixtures MoE. 

\subsection{Video-level Multi-modal Fusion}

\begin{figure}[t]
\centering
\includegraphics[height=4.8cm]{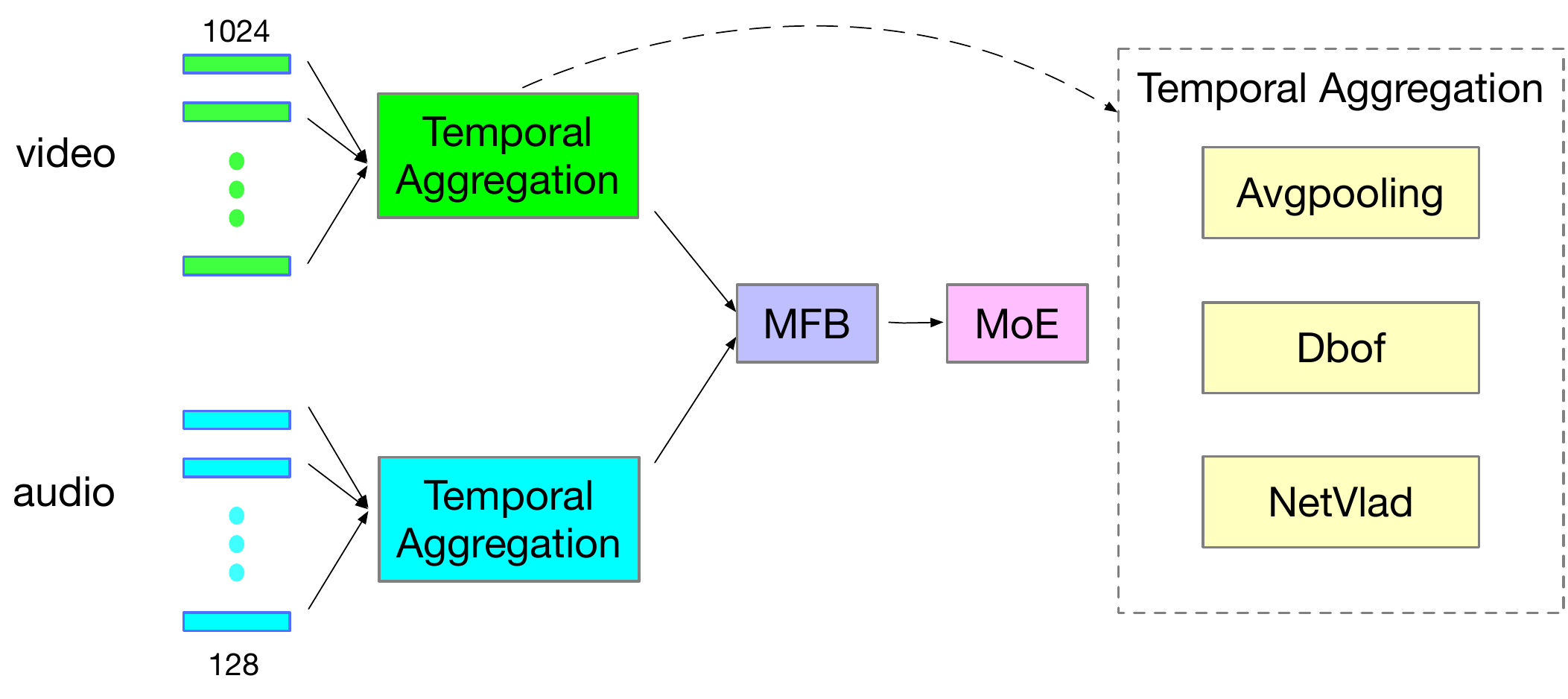}
\caption{The overall architecture of our MFB augmented video classification system.}
\label{fig:video_level_mfb}
\end{figure}
In this section we will illustate that MFB module can be a plug-and-play layer to fuse aggregated visual and audio features.  Fig.~\ref{fig:video_level_mfb} shows the overall video-level fusion architecture. It mainly contains three parts. Firstly, the pre-extracted visual features $\textbf{L}$ and audio features $\textbf{A}$ are fed into  two temporal aggregation modules separately. %, the temporal aggregation module has a two-stream architecture treating visual and audio separately. 
Each module outputs a single compact video-level representation and can be any one of the mentioned three aggregating mechanisms shown in the right side of figure. Next, MFB module fuse aggregated visual and audio features into a fixed-length representation. Finally, the classification module takes the resulting compact representation as input and outputs confidence scores of each semantic label. Following~\cite{Abu2016YouTube}, we adopt Mixture-of-Experts~\cite{Jordan1994Hierarchical} as our classifier. The Mixture of Experts~\cite{Jordan1994Hierarchical}  classifier layer consists of  $m$ ``expert networks'' which take the global multimodal representation $f$ as input and estimate a distribution over $c$ classes. The final prediction $\textbf{d}$ is defined as 
\begin{equation}
    \textbf{d} = \sum_{i=1}^{m}\mathrm{softmax}(\textbf{g}_i) \odot sigmoid(\textbf{e}_i),
\end{equation}
\begin{equation}
    \textbf{g}_{i} = \textbf{fW}_{g, i} + \lambda ||\textbf{W}_{g,i}||_{2},
\end{equation}
\begin{equation}
    \textbf{e}_{i} = \textbf{fW}_{e, i} + \lambda ||\textbf{W}_{e,i}||_{2},
\end{equation}
where $\textbf{W}_{g,i}, \textbf{W}_{e,i}, i \in \{1, ..., m\}$ are trainable paramters and $O \in \mathbb{R}^{c}$. $\lambda$ is the L2 penalty with the default value 1e-6. All our models are trained with 2-mixtures MoE. 

\section{Experiments}

\subsection{Implementation details}
We implement our model based on Google starter code\footnotemark[1]. Each training is performed on a single V100 (16Gb) GPU. All our models are trained using Adam optimizer~\cite{kingma2014adam} with an initial learning rate set to 0.0002. The mini-batch size is set to 128. We found that cross entropy classification loss works well for maximizing the Global Average Precision (GAP).
All model are trained with \num{250000} steps. In order to observe timely model prediction, we evaluate the model on a subset of validate set every \num{10000} training steps. For the cluster-based pooling, the cluster size $K$ is set to 8 for NetVLAD and \num{2000} for DBoF. To have a fair comparison, 300 frames are sampled before aggregation. In addition, the dropout rate of MFB module is set to 0.1 in all our expriements.

\footnotetext[1]{https://github.com/google/youtube-8m}

\subsection{Datasets and evaluation metrics}

We conduct experiments on the recently updated Youtube-8M v2 dataset with improved machine-generated labels and  higher-quality videos. It contains a total of 6.1 million videos, \num{3862} classes, 3 labels per video averagely. Visual and audio features are pre-extracted per frame. Visual features are obtained by  Google Inception convolutional neural network pretrained on ImageNet~\cite{deng2009imagenet}, followed by PCA-compression to generate a vector with 1024 dimensions. The audio features are extracted from a VGG-inspired acoustic model described in ~\cite{vgg_audio}. In the official split, training, validataion and test have equal \num{3844} tfrecord shards. In practice, we use \num{3844} training shards and \num{3000} validation shards for training. We randomly select 200 shards from the rest of 844 validation shards (around \num{243337} videos) to evaluate our model every \num{10000} training steps. %We test our model on the 3844 test shards. 
Results are evaluated using the Global Average Precision (GAP) metric at top 20 as used in the Youtube-8M Kaggle competition. %We have noticed that the performance on our validation set is comparable ($\pm 0.3\%$) to the test performance evaluated on the Kaggle platform, so we only report the results on the validation set.

\setlength{\tabcolsep}{4pt}
\begin{table}[t]
\begin{center}
\caption{Comparision study on Avgpooling feature}
\label{table:avgpooling}
\begin{tabular}{lll}
\hline\noalign{\smallskip}
Model & GAP\\ 
\noalign{\smallskip}
\hline
\noalign{\smallskip}
Avgpooling + Audio Only  & 38.1 \\
Avgpooling + Video Only  & 69.6 \\
Avgpooling + Concatenation  & 74.2 \\
Avgpooling + FC + Concatenation & 81.8 \\
Avgpooling + MFB & \textbf{83.3} \\
\hline
\end{tabular}
\end{center}
\end{table}
\setlength{\tabcolsep}{1.4pt}

\setlength{\tabcolsep}{4pt}
\begin{table}[t]
\begin{center}
\caption{Comparision study on  NetVLAD feature}
\label{table:NetVlad}
\begin{tabular}{lll}
\hline\noalign{\smallskip}
Model & GAP\\ 
\noalign{\smallskip}
\hline
\noalign{\smallskip}
NetVLAD + Audio Only  & 50.7 \\
NetVLAD + Video Only  & 82.3 \\
NetVLAD + Concatenation  & 85.0 \\
NetVLAD + FC + Concatenation & 84.6 \\
NetVLAD + MFB & \textbf{85.5} \\
\hline
\end{tabular}
\end{center}
\end{table}
\setlength{\tabcolsep}{1.4pt}

\setlength{\tabcolsep}{4pt}
\begin{table}[]
\begin{center}
\caption{Comparision study on DBoF feature}
\label{table:DBoF}
\begin{tabular}{lll}
\hline\noalign{\smallskip}
Model & GAP\\ 
\noalign{\smallskip}
\hline
\noalign{\smallskip}
DBoF + Audio Only & 48.9 \\
DBoF + Video Only & 81.8 \\
DBoF + Concatenation  & 84.0 \\
DBoF + FC + Concatenation & 84.1 \\
DBoF + MFB & \textbf{85.9} \\
\hline
\end{tabular}
\end{center}
\end{table}
\setlength{\tabcolsep}{1.4pt}

\subsection{Results}

In this section, we verify the effectiveness of MFB module by comparing its performance on the validation set with the simple concatenation fusion. We also conduct two comparative tests with single-modality input (only video or audio). To prove that the improvement of performance does not come from increasing parameters, we add another comparison with the same number of parameters as MFB. Specifically, the temporal aggregated video and audio representations are first projected using a fully connected layer respectively and then the projected video and audio vectors are concatenated to feed into the MoE classifier (For convenience, we call it FC Concatenation module later). The fully connected layers have the same parameter settings with those in MFB module. The superior GAP performance of MFB module on three temporal aggregation models is shown as follows. 

The detailed results of MFB with Avgpooling features are shown in  Tab~\ref{table:avgpooling}. Firstly, the GAP performance of two modal fusion is far superior to single modality input (Video Only or Audio Only). In the NetVLAD and DBof video features, we can draw the same conclusion. Secondly, we can observe a significant increase in performance with the MFB module, which achieves a 9.1\% higher GAP compared with the concatenation fusion baseline. Even if the concatenation is augmented with the same number of parameters as MFB,  there is still a $1.5\%$ gap. The main reason is probably that the simple fusion can not leverage high-order information across modalities. % the MFB module also improves the GAP from 81.8\% to 83.3\%. 

In terms of NetVLAD video features, the MFB module improves the GAP from 85.0\% to 85.5\% compared to the concatenation module as shown in Tab~\ref{table:NetVlad}. However surprisingly, adding fully connected layers %in FC Concatenation module
performs worse, indicating that  NetVLAD has been a quite good temporal model for single modal data aggregation. In some sense, increasing parameters will lead to overfitting . Therefore, it also proves that  MFB contributes to the performance boost. For DBoF, the results are consistent with Avgpooling and NetVLAD, MFB module achieves the best GAP of 85.9\%, around 1.8\% higher than another two methods. We conclude that MFB  encourages abundant cross-modal interactions and thus reduce the ambiguity of each modal data. 

\begin{figure}[t]
\centering
\includegraphics[width=\linewidth]{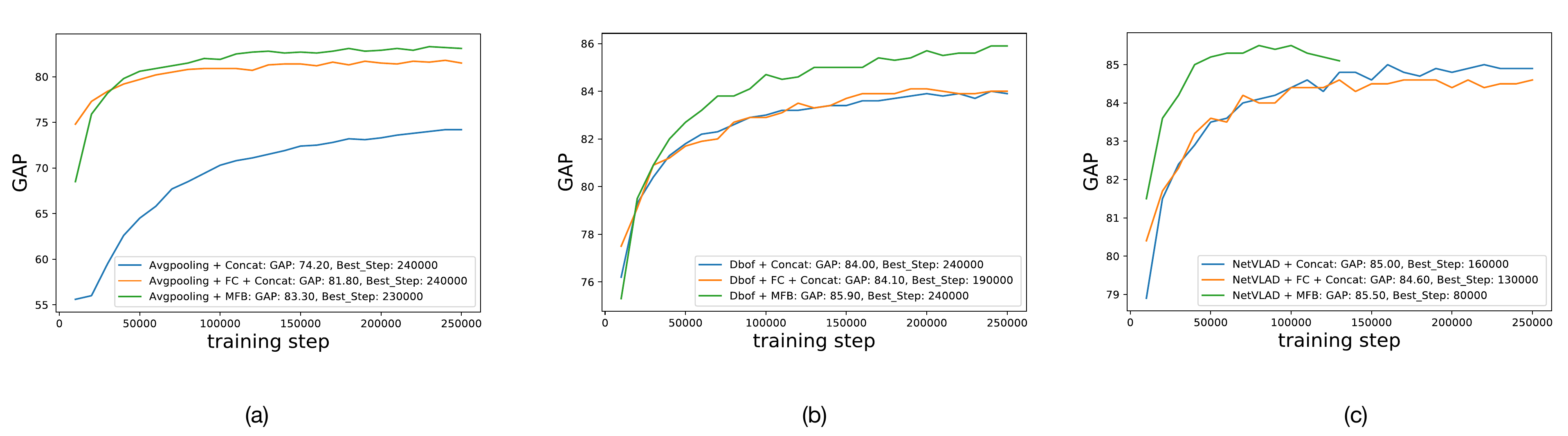}
\caption{(a) The GAP Performance of Avgpooling feature with different fusion modules (Concantenation, FC Concantenation, MFB). (b) The GAP Performance of DBoF feature. (c) The GAP Performance of NetVLAD feature.}
\label{fig:training}
\end{figure}

In order to give an intuitive observation on the advantage of MFB over simple fusion baselines, we illustrate the training processes of three fusion modules in Fig.~\ref{fig:training}, which shows the GAP performance on validation dataset as the training iteration increases. It is worthy noting that the experiment with the MFB module and  NetVLAD features is early stopped at around \num{13000} steps due to overfitting. Obviously, MFB module can not only increase the capability of video and audio fusion but also speed-up training.

\section{Conclusions}

In this paper, we first apply the multimodal factorized blinear pooling into large-scale video classification task. To validate its effectiveness and robustness, we experiment on three kinds of video-level features obtained by Avgpooling, NetVLAD and DBoF. We conduct experiments on large-scale video classification benchmark Youtube-8M. Experimental results demonstrate that the carefully designed multimodal factorized bilinear pooling can achieve significantly better results than the popular fusion concatenation operator. Our future work mainly lies on directly combining multimodal factorized bilinear pooling with multimodal frame-level data.

% \bibliographystyle{splncs}
% \bibliography{eccv}

\begin{thebibliography}{10}

\bibitem{Krizhevsky2012ImageNet}
Krizhevsky, A., Sutskever, I., Hinton, G.E.:
\newblock Imagenet classification with deep convolutional neural networks.
\newblock In: International Conference on Neural Information Processing
  Systems. (2012)  1097--1105

\bibitem{He2016Deep}
He, K., Zhang, X., Ren, S., Sun, J.:
\newblock Deep residual learning for image recognition.
\newblock In: IEEE Conference on Computer Vision and Pattern Recognition.
  (2016)  770--778

\bibitem{Szegedy2016Rethinking}
Szegedy, C., Vanhoucke, V., Ioffe, S., Shlens, J., Wojna, Z.:
\newblock Rethinking the inception architecture for computer vision.
\newblock In: IEEE Conference on Computer Vision and Pattern Recognition.
  (2016)  2818--2826

\bibitem{simonyan2014two}
Simonyan, K., Zisserman, A.:
\newblock Two-stream convolutional networks for action recognition in videos.
\newblock In: Advances in neural information processing systems. (2014)
  568--576

\bibitem{wang2016temporal}
Wang, L., Xiong, Y., Wang, Z., Qiao, Y., Lin, D., Tang, X., Van~Gool, L.:
\newblock Temporal segment networks: Towards good practices for deep action
  recognition.
\newblock In: European Conference on Computer Vision, Springer (2016)  20--36

\bibitem{Girdhar2017ActionVLAD}
Girdhar, R., Ramanan, D., Gupta, A., Sivic, J., Russell, B.:
\newblock Actionvlad: Learning spatio-temporal aggregation for action
  classification.
\newblock (2017)  3165--3174

\bibitem{hinton2012deep}
Hinton, G., Deng, L., Yu, D., Dahl, G.E., Mohamed, A.r., Jaitly, N., Senior,
  A., Vanhoucke, V., Nguyen, P., Sainath, T.N.,  et~al.:
\newblock Deep neural networks for acoustic modeling in speech recognition: The
  shared views of four research groups.
\newblock IEEE Signal processing magazine \textbf{29}(6) (2012)  82--97

\bibitem{vgg_audio}
Hershey, S., Chaudhuri, S., Ellis, D.P.W., Gemmeke, J.F., Jansen, A., Moore,
  C., Plakal, M., Platt, D., Saurous, R.A., Seybold, B., Slaney, M., Weiss, R.,
  Wilson, K.:
\newblock Cnn architectures for large-scale audio classification.
\newblock In: International Conference on Acoustics, Speech and Signal
  Processing (ICASSP).
\newblock (2017)

\bibitem{Abu2016YouTube}
Abu-El-Haija, S., Kothari, N., Lee, J., Natsev, P., Toderici, G., Varadarajan,
  B., Vijayanarasimhan, S.:
\newblock Youtube-8m: A large-scale video classification benchmark.
\newblock (2016)

\bibitem{hochreiter1997long}
Hochreiter, S., Schmidhuber, J.:
\newblock Long short-term memory.
\newblock Neural computation \textbf{9}(8) (1997)  1735--1780

\bibitem{cho2014properties}
Cho, K., Van~Merri{\"e}nboer, B., Bahdanau, D., Bengio, Y.:
\newblock On the properties of neural machine translation: Encoder-decoder
  approaches.
\newblock arXiv preprint arXiv:1409.1259 (2014)

\bibitem{Gao2015Compact}
Gao, Y., Beijbom, O., Zhang, N., Darrell, T.:
\newblock Compact bilinear pooling.
\newblock (2015)

\bibitem{kim2016hadamard}
Kim, J.H., On, K.W., Lim, W., Kim, J., Ha, J.W., Zhang, B.T.:
\newblock Hadamard product for low-rank bilinear pooling.
\newblock arXiv preprint arXiv:1610.04325 (2016)

\bibitem{yu2017mfb}
Yu, Z., Yu, J., Fan, J., Tao, D.:
\newblock Multi-modal factorized bilinear pooling with co-attention learning
  for visual question answering.
\newblock IEEE International Conference on Computer Vision (ICCV) (2017)
  1839--1848

\bibitem{arandjelovic2016netvlad}
Arandjelovic, R., Gronat, P., Torii, A., Pajdla, T., Sivic, J.:
\newblock Netvlad: Cnn architecture for weakly supervised place recognition.
\newblock In: Proceedings of the IEEE Conference on Computer Vision and Pattern
  Recognition. (2016)  5297--5307

\bibitem{deng2009imagenet}
Deng, J., Dong, W., Socher, R., Li, L.J., Li, K., Fei-Fei, L.:
\newblock Imagenet: A large-scale hierarchical image database.
\newblock In: Computer Vision and Pattern Recognition, 2009. CVPR 2009. IEEE
  Conference on, Ieee (2009)  248--255

\bibitem{caba2015activitynet}
Caba~Heilbron, F., Escorcia, V., Ghanem, B., Carlos~Niebles, J.:
\newblock Activitynet: A large-scale video benchmark for human activity
  understanding.
\newblock In: Proceedings of the IEEE Conference on Computer Vision and Pattern
  Recognition. (2015)  961--970

\bibitem{tran2018closer}
Tran, D., Wang, H., Torresani, L., Ray, J., LeCun, Y., Paluri, M.:
\newblock A closer look at spatiotemporal convolutions for action recognition.
\newblock In: Proceedings of the IEEE Conference on Computer Vision and Pattern
  Recognition. (2018)  6450--6459

\bibitem{qiu2017learning}
Qiu, Z., Yao, T., Mei, T.:
\newblock Learning spatio-temporal representation with pseudo-3d residual
  networks.
\newblock In: ICCV. (2017)

\bibitem{carreira2017quo}
Carreira, J., Zisserman, A.:
\newblock Quo vadis, action recognition? a new model and the kinetics dataset.
\newblock In: Computer Vision and Pattern Recognition (CVPR), 2017 IEEE
  Conference on, IEEE (2017)  4724--4733

\bibitem{J2010Aggregating}
Jégou, H., Douze, M., Schmid, C., Pérez, P.:
\newblock Aggregating local descriptors into a compact image representation.
\newblock In: Computer Vision and Pattern Recognition. (2010)  3304--3311

\bibitem{Perronnin2007Fisher}
Perronnin, F., Dance, C.:
\newblock Fisher kernels on visual vocabularies for image categorization.
\newblock In: IEEE Conference on Computer Vision and Pattern Recognition.
  (2007)  1--8

\bibitem{ben2017mutan}
Ben-Younes, H., Cadene, R., Cord, M., Thome, N.:
\newblock Mutan: Multimodal tucker fusion for visual question answering.
\newblock In: Proc. IEEE Int. Conf. Comp. Vis. Volume~3. (2017)

\bibitem{Miech2017Learnable}
Miech, A., Laptev, I., Sivic, J.:
\newblock Learnable pooling with context gating for video classification.
\newblock (2017)

\bibitem{Pirsiavash2009Bilinear}
Pirsiavash, H., Ramanan, D., Fowlkes, C.:
\newblock Bilinear classifiers for visual recognition.
\newblock In: International Conference on Neural Information Processing
  Systems. (2009)  1482--1490

\bibitem{Ng2015Beyond}
Ng, Y.H., Hausknecht, M., Vijayanarasimhan, S., Vinyals, O., Monga, R.,
  Toderici, G.:
\newblock Beyond short snippets: Deep networks for video classification.
\newblock \textbf{16}(4) (2015)  4694--4702

\bibitem{Wang2009Evaluation}
Wang, H., Ullah, M.M., Kläser, A., Laptev, I., Schmid, C.:
\newblock Evaluation of local spatio-temporal features for action recognition.
\newblock In: British Machine Vision Conference, BMVC 2009, London, UK,
  September 7-10, 2009. Proceedings. (2009)

\bibitem{Jordan1994Hierarchical}
Jordan, M.I., Jacobs, R.A.:
\newblock Hierarchical mixtures of experts and the EM algorithm.
\newblock Springer London (1994)

\bibitem{kingma2014adam}
Kingma, D.P., Ba, J.:
\newblock Adam: A method for stochastic optimization.
\newblock arXiv preprint arXiv:1412.6980 (2014)

\end{thebibliography}

\end{document}